\documentclass[11pt,a4paper]{article}
\usepackage[hyperref]{naaclhlt2018}
\usepackage{times}
\usepackage{latexsym}
\usepackage{multirow}
\usepackage{amsmath,amsfonts,amssymb}
\usepackage{graphicx} 
\usepackage{url}

\aclfinalcopy 

\title{UMDSub at SemEval-2018 Task 2: Multilingual Emoji Prediction\\ 
Multi-channel Convolutional Neural Network on Subword Embedding}

\author{
  Zhenduo Wang \and Ted Pedersen\\
  Department of Computer Science\\
  University of Minnesota \\
  Duluth, MN 55812, USA \\
  {\tt\{wang7211,tpederse\}@d.umn.edu}
}

\date{}

\begin{document}
\maketitle

\begin{abstract}
This paper describes the UMDSub system that participated in Task 2 of SemEval-2018. 
We developed a system that predicts an emoji given the raw text in a English tweet.
The system is a Multi-channel Convolutional Neural Network based on subword embeddings for 
the representation of tweets. This model improves on character or word based methods by about 2\%.
Our system placed 21st of 48 participating systems in the official evaluation.
\end{abstract}

\section{Introduction}

People use emojis on social media with text in order to express their emotions or reveal meanings hidden by 
figurative language. However, understanding the semantic relationship between texts and emojis is 
a challenging task because emojis often have different interpretations that depend on the reader. 
\cite{semeval2018task2} shows that a carefully designed machine learning system can 
perform better than humans on an emoji prediction task. The system which gives the best predictions 
in that paper is a long short term memory (LSTM) network with pre-trained character embeddings as 
the text representation method. 

\cite{semeval2018task2} provides a new task in the context of Twitter and describes several 
powerful models in text sequence modeling and classification. Our work explores other possible 
configurations for the system, including subword embedding as a text representation method and multi-channel Convolutional Neural Networks as a feature extraction method. We conduct experiments on the Semeval-2018 Task 2 Multilingual Emoji Prediction data set and compare our results with other methods. Our results show that subword embedding based Convolutional Neural Network system is effective. It improves on character embedding by 2.1\% and word embedding by 1.8\%. 

\section{Task Description}

SemEval--2018 Task 2 has two subtasks which are similar but in different languages. 
Subtask A is emoji prediction on English tweets, while subtask B is on Spanish. Both tasks are to predict 1 out of 20 emojis (19 in Spanish subtask) given only the text of a tweet. There are $\sim$500K tweets in the English training set and $\sim$100K tweets in the Spanish training set. The organizers use the macro F-score to evaluate the performance of the systems.

This is by definition a text classification task. It is more challenging than tasks such as sentiment analysis or authorship identification because it is on Twitter data which is extremely large and changes frequently. We build UMDSub for the English subtask and conduct several experiments with different settings.

\section{System Description}

UMDSub is a multi-channel Convolutional Neural Network based on subword embedding as the text representation. We use byte-pair-encoding for subword segmentation.

\subsection{Word Segmentation}

The goal of the word segmentation step is to break the words into subwords in order to have better representations for tweet texts. 

Existing text representation methods mostly work on whole word level \citep{word2vec}\citep{Glove} or character level \citep{char-CNN} \citep{char-CRNN}. In these works, character embedding as the basis for feature extraction were shown to be more effective on text classification tasks. Also there are works such as \citep{fasttext} which try to enrich word representations with subword information and improve upon the original word embedding methods. We believe that subword is a intriguing level for text representation. However, the method used to generate subwords in \citep{fasttext} is to divide words into fixed length n-grams, which fails to use knowledge of word morphology. We believe that \textit{Language is never, ever, ever, random} \cite{kilgarriff2005language}. Subwords as character n-grams could be detected by measuring their frequencies.

We detect subwords with the byte-pair-encoding (BPE) algorithm \citep{BPE}, which is a frequency based text compression algorithm. The original algorithm first splits the text into bytes (characters). Then it ranks all the byte pairs (which are simply bigrams) according to their frequencies. The most frequent byte pair is then joined together and encoded by a single byte. By repeating this simple process, the original text will be compressed. In our system, we use BPE as a word segmentation algorithm. We join characters together by frequency but do not replace them with new symbols. This method was also shown to be successful in \citep{BPENMT}. 

For example, suppose the text to be segmented is
\begin{center}
\textit{$S_0$ = workers work in workshop.}
\end{center}
First, we split $S_0$ into character sequence.
\begin{center}
\textit{$S_1$ = w o r k e r s w o r k i n w o r k s h o p.}
\end{center}
Then in the first iteration, we choose the most frequent bigram \textit{wo} and join them together.
\begin{center}
\textit{$S_2$ = \underline{wo} r k e r s \underline{wo} r k i n \underline{wo} r k s h o p.}
\end{center}
In the second iteration, we choose the most frequent bigram \textit{\underline{wo}r} and join them together.
\begin{center}
\textit{$S_3$ = \underline{wor} k e r s \underline{wor} k i n \underline{wor} k s h o p.}
\end{center}
Similarly, the next bigram should be \textit{\underline{wor}k}.
\begin{center}
\textit{$S_4$ = \underline{work} e r s \underline{work} i n \underline{work} s h o p.}
\end{center}
The text is represented by a subword sequence after the algorithm finishes in $N$ iterations, and will look like this :
\begin{center}
\textit{$S_N$ = \underline{work} \underline{er} s \underline{work} \underline{in} \underline{work} \underline{shop}.}
\end{center}

\subsection{Subword Embedding Layer}

Embeddings can be categorized into two types based on how they are trained, saved (or static) and reused (or dynamic). Static embeddings are separately trained from large open corpus (such as Wikipedia) and saved for reuse. Word2vec and GloVe models are two such examples. Dynamic embeddings are jointly trained with other parts of specific systems. The Twitter corpus is not like a standard corpus, since it is extremely large and it changes frequently. We choose to use a dynamic embedding strategy since it may fit better for our system. 

After word segmentation the text is made up of subword sequences. In order to use subword embedding to represent the text, we first represent each subword type with a one-hot vector. The one-hot vector for the $i$th subword in vocabulary is a sparse binary vector $o_{i}$ which has 1 as the $i$th element and 0 for all others. This step results in a representation similar to a vector space model. We put all the subwords in the subword one-hot hyperspace where each unique subword type owns one dimension. The one-hot embedding is so sparse and it suffers from the Dimension Disaster. Hence we project it to a smaller hyperspace by multiplying the one-hot embedding with a projection weighting matrix $W\in\mathbb{R}^{d\times|V|}$, where $d$ is the dimension for the target embedding hyperspace and $|V|$ is the subword vocabulary size. Now each subword is represented by a dense vector $s_i=Wo_i$, and the tweet text $T$ with length of $L$ is represented by a sequence of subword embedding vectors $T=(s_1,s_2,...,s_L)$.

\subsection{Multi-channel Convolutional Layer}

As we mentioned in Section 2, this is a text classification task. A key step in any classification system which uses the feature-classifier scheme is to choose indicative features. In NLP tasks, n-grams are commonly used features because they capture word collocations as found in text. But this may not be effective if we have no sense of what kind(s) of n-grams could be useful for a certain problem. Recently, a trend \citep{char-CNN} \citep{char-CRNN} for solving text classification tasks is to break the text into the smallest units and build a representation for them. Then this approach can use a  complex neural network such as Convolutional Neural Network (CNN) or recurrent neural network (RNN) to find abstract features upon the representations. This system configuration takes advantage of the ability of complex neural networks to extract features. \citep{kim2016char-CNN} tries to explain the mechanism of character-entry-CNN by showing that the features extracted by CNN are specific n-grams. We believe this scheme could be very useful for finding features for classification when we have little knowledge of the corpus, as is often the case for Twitter.

In our system, we use a multi-channel CNN layer for feature extraction. Since tweets are generally short, higher-level features for classification may not exist. Instead of adding depth to our network, we add diversity to the kernel sizes in order to keep as much n-gram information as possible. A similar model is used in \cite{multi-channelAA} \cite{cnnaa}. 

The multi-channel convolution layer consists of three parallel convolution units. They have a similar structure but different convolutional kernel sizes. Each convolution unit consists of two steps, the convolution and pooling. The convolution step is to calculate the convolutions of the resulting vector sequence from the subword embedding vector sequence $T=(s_1,s_2,...,s_L)$ and convolution kernels $k_{1:M}\in \mathbb{R}^{d,M\times r}$, where $M$ is the kernel number, $r$ is the kernel size. The kernel size represents the context window of feature extraction and the kernel number represents the number of patterns.

\[ 
f_{l=1:L}^{m'}=\sigma(k_m\ast[s_{l-r/2+1},...,s_l,..,s_{l+r/2}])
\]
where $\sigma$ is the activation function.

Then the pooling step is to trim the resulting sequence $F'$ by leaving only the maximum in every $r$ consecutive $f_l'$s.
\[ 
f_{q=1:L/r}^{m}=\max\{f_{l=(q-1)\times r+1}^{m'},...,f_{l=q\times r}^{m'}\}
\]
Then the convolution unit outputs features extracted by each kernel. The output of can be seen as all the features extracted within a certain size of context window.
\[ 
F=\left[
  f_{1}^{1},...,f_{1}^{M}
  \right]
\]

Each CNN unit works as above. We concatenate all the three feature maps and flatten the resulting matrix to one single vector. This single vector is later used for classification.

\subsection{Classification Layer} 

We make two assumptions of how texts affect the usage of emojis, and how emojis are chosen to express emotions as an auxiliary symbol for the text.

\begin{enumerate}
\item Emojis are chosen and used to indicate the emotion of the whole tweet or the emotion of the tweeter at the moment, independent of position.
\item Emojis are tied to its context only, enhancing or revealing the true underlying meaning. This can vary depending on the position or content of the tweet.
\end{enumerate}

In case 2, a ``verbose" tweet which contains several sentences may map to different emojis at different positions. Given this nature of how emojis are used in tweets, the task could have been very different from traditional text classification since only a part of the text decides the classification output. Fortunately, most the tweets are chosen so that they contain only one sentence and the dataset does not provide the position information of the removed emoji, which makes the task easier. Hence we assume all the emojis are used as in case 1 and we will use the whole tweet to do prediction.

From the convolutional layer we get output of the features extracted from the whole tweet. We first reshape the features matrix into a single vector, then we feed it to a logistic regression layer for the final classification. The layer consists of a fully connected network layer and a softmax function as activation function. It takes the the one dimensional feature vector generated from convolutional layer as input and outputs a distribution over the 20 emojis denoting their probabilities of being used in the original tweet. Hence this layer can be represented by:
\[ P[y=k|\vec{X}]= \dfrac{\exp(W_k^\top \vec{X}+b_k)}{ \sum_{i=1}^{20} \exp(W_i^\top\vec{X}+b_i)}\]

Figure 1 is a summary of our system.
	\begin{figure}[thpb]
    	\centering
		\includegraphics[width=6cm]{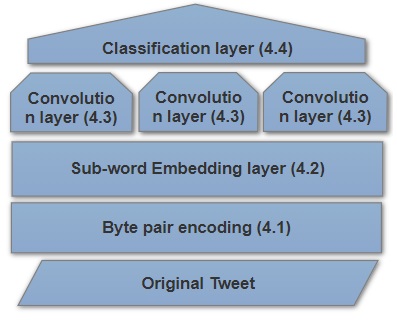}
		\caption{Diagram of UMDSub.}
		\label{figurelabel}
	\end{figure}

\section{Parameter Settings}

We use the training and test tweets collectively for the byte pair encoding 
algorithm with 2,000 iteration. Each iteration of BPE algorithm generates a new character 
n-gram, resulting in a vocabulary size around 2,000. Then we embed 
each subword with a vector with dimension $d=64$. In our concatenated Convolutional 
Neural Network, each unit has $M=128$ filters with filter sizes and max pooling size $r=\{3,4,5\}$.  
For comparison, we also build character based system containing not only the ascii 
characters with vocabulary size $|V|=200$ and embedding dimension $d=16$ and word based system with vocabulary size $|V|=10,000$ 
and embedding dimension $d=256$. We use the same network structure for all different text representation methods. 
We train the network for 50 epochs. Table 1 summarizes the parameter settings used.

\begin{center}
\begin{table}
\begin{tabular}{l*{5}{c}r}
\hline
Embedding 	  & $|V|$ & $d$ & $M$ & $r$ &$\#$param\\
\hline
Word            & 10K & 256 & \multirow{3}{*}{128} & \multirow{3}{*}{3, 4, 5}  & 4.16M \\
subword 		& 2K & 64 &  &  & 0.63M\\
Character       & 200 & 16 &  &   & 0.13M\\
\hline 
\end{tabular}
\caption{Network parameter settings.}
\end{table}
\end{center}

\section{Experimental Results}

Table 2 shows our experimental results  and
reveals that our methods have a 
preference towards precision. In the task evaluation, 
UMDSub attained precision of .330 which was 
9th of 48 systems, and recall of .267 which was 20th.
The overall F-score was .260 and placed 21st. 
This emphasis on precision can be seen in that our system was particularly accurate
in predicting the most frequent emoji in the training and test data (the red heart), 
achieving an accuracy of .854
which was 4th among the 48 participating systems.

Our submission for the evaluation phase (sub-CNN (E)) was produced by a 4-layer single channel 
Convolutional Neural Network. After we changed it to the post-evaluation version of 
our multi-channel network (sub-CNN (P)), we saw a significant improvement where the F-score 
rose to .301 which would have placed 10th in the evaluation. 
The reasons for this success have to do with the nature of tweets, which
are difficult to represent in a deep network given their very limited content.
A Multi-channel CNN captures 
more information using various kernels in a single layer and so the content of
short and somewhat noisy tweets is well represented.

\begin{center}
\begin{table}
\begin{tabular}{l*{4}{c}r}
\hline
System config 	  & F1  &  Precision  & Recall\\
\hline
sub-CNN (E) & .260 & .330 & .267\\
Word-CNN (P)& .283 & .355 & .285\\
sub-CNN (P) & \bf.301 & .352 & \bf.302\\
Char-CNN (P)& .289 & \bf.382  & .291 \\
\hline
BOW & .29 & .32 & .34  \\
Word-LSTM & .33 & .35 & .36  \\
Word-LSTM + P & .32& .34 & .36   \\
Char-LSTM & .32 & .36 & .37   \\
Char-LSTM + P & \bf{.34} & \bf.42 & \bf.39   \\
\hline 
\end{tabular}
\caption{Experimental results (both evaluation (E) and post-evaluation (P) phase). -LSTM refers to the models used in \cite{semeval2018task2}, with results from this paper. +P refers to pre-trained embeddings.}
\end{table}
\end{center}

\section{Conclusion}

Our work shows that subword embedding is an effective method of text representation 
for the emoji prediction task. Under the multi-channel CNN framework, it improves 
word embedding by 1.8\% and character embedding by 2.1\% while maintaining a modest computational cost. 

We try to explain the reason of our results with the success of byte pair encoding 
(BPE) segmentation. We check the segmented text after byte pair encoding algorithm 
and we find a very representative example: 

\textbf{Before BPE:} Playing the drums on RockBand made it look much easier than it is.

\textbf{After BPE:} Pl ay ing the dr um s on Rock B and made it look much ea si er than it is.

We observe that some of the segmentations correspond with word morphology, others not. 
This is because the segmentation is based on frequency and not a knowledge of morphology.
Based on these observations, we consider the BPE algorithm a method that dynamically 
decides which n-grams are most frequent in a corpus and thus deserving of a unique 
representation that is included in the vocabulary. The vocabulary generated by 
BPE is a mixture of characters, roots, affixes, whole words and even random n-grams, 
chosen as such to make the representation of text statistically efficient. The 
advantage of such a system is that it does not consider Twitter language as a 
compound made of characters or atomic words. Therefore it does not limit text 
representation on only one single level, making the later feature extraction more efficient.

\nocite{*}

\bibliography{semeval}
\bibliographystyle{acl_natbib}

\end{document}